\documentclass[journal, a4paper]{IEEEtran}

\usepackage[utf8x]{inputenc}
\usepackage{amsmath}
\usepackage{amssymb}
\usepackage{array}
\usepackage{epsfig}
\usepackage{chngcntr}
\usepackage{graphicx}
\usepackage{subfigure}

\usepackage[noadjust]{cite}

\usepackage{todonotes}
\usepackage{leftidx}
\usepackage{comment}
\usepackage{url}
\usepackage{tabularx}

\usepackage[colorlinks = false,
linkcolor = black,
urlcolor  = blue,
citecolor = black,
anchorcolor = blue,
hidelinks = true]{hyperref}

\usepackage[lined,ruled]{algorithm2e}

\usepackage{multirow}

\usepackage{amsthm}

\theoremstyle{plain}

\theoremstyle{definition}

\makeatletter
\DeclareRobustCommand{\append}[2]{%
	\mathpalette\app@nd{{#1}{#2}}#1%
}
\newcommand{\app@nd}[2]{\app@@nd#1#2}
\newcommand{\app@@nd}[3]{%
	\mbox{\scriptspace\trylength\m@th$#1{\vphantom{#2}}^{#3}$}%
}
\makeatother

\usepackage{epstopdf}
   \graphicspath{{./figs/}}
   \DeclareGraphicsExtensions{.pdf,.jpg,.png,.eps}

\IEEEoverridecommandlockouts      
\hypersetup{
	bookmarks=true,         
	unicode=false,          
	pdftoolbar=true,        
	pdfmenubar=true,        
	pdffitwindow=false,     
	pdftitle={MultiVehicleCooperativeLocalizationVx},    
	pdfauthor={Dr. Xiaotong Shen},     
	pdfsubject={paper},   
	pdfcreator={Dr. Xiaotong Shen},   
	pdfproducer={},  
	pdfnewwindow=true,      
	colorlinks=false,       
	linkcolor=white,          
	citecolor=white,        
	filecolor=white,      
	urlcolor=white           
}

\title{
A General Framework for Multi-vehicle Cooperative Localization Using Pose Graph
}

\author{
Xiaotong Shen,
Hans Andersen,
Wei Kang Leong,
Hai Xun Kong,
Marcelo H. Ang Jr.,
and
Daniela Rus
\thanks{X. Shen, W. K. Leong and H. X. Kong are with Singapore-MIT Alliance for Research and Technology (SMART), Singapore, {xiaotong,weikang,haixun@smart.mit.edu}.}
\thanks{H. Andersen and M. H. Ang Jr. are with the Department of Mechanical Engineering, National University of Singapore, Kent Ridge, Singapore, {mpeangh@nus.edu.sg,hans.andersen@u.nus.edu}.}
\thanks{D. Rus is with Massachusetts Institute of Technology, Cambridge, MA., USA, rus@csail.mit.edu.}
\thanks{This research was supported by the Future Urban Mobility project of the Singapore-MIT Alliance for Research and Technology (SMART) Center, with funding from Singapore's National Research Foundation.}
}

\begin{document}

\maketitle

\begin{abstract}
When a vehicle observes another one, the two vehicles' poses are correlated by this spatial relative observation, which can be used in cooperative localization for further increasing localization accuracy and precision. To use spatial relative observations, we propose to add them into a pose graph for optimal pose estimation. Before adding them, we need to know the identities of the observed vehicles. The vehicle identification is formulated as a linear assignment problem, which can be solved efficiently. By using pose graph techniques and the start-of-the-art factor composition/decomposition method, our cooperative localization algorithm is robust against communication delay, packet loss, and out-of-sequence packet reception. We demonstrate the usability of our framework and effectiveness of our algorithm through both simulations and real-world experiments using three vehicles on the road. 
\end{abstract}


\section{Introduction}

Information sharing among vehicles has great impact on improving driving safety and smoothening traffic flow \cite{Multi2012Li,Motion2013Liu,Vehicle2015Kim}. With sensing information being shared over vehicle-to-vehicle (V2V) communication networks \cite{Multi2015Shen}, the perception range can be extended beyond line-of-sight and field-of-view \cite{Cooperative2013Kim,Cooperative2013Li,Multivehicle2015Kim}. In order to fuse the shared information and obtain not only augmented but also consistent observations, the uncertainties of vehicles' poses should be minimized to a reasonable level \cite{Scalable2016Shen}. Cooperative localization can improve pose estimation precision and accuracy by utilizing relative observations as correlations between vehicle poses, so that the individual perception results would be aligned and consistent when projecting the shared information on a global map.

Communication issues, such as delay, packet loss, and out-of-sequence packet reception, pose big challenges to cooperative localization. Among them, packet loss has great influence on the estimation uncertainty \cite{sinopoli2004kalman,Bounds2015Shen}. Especially for on-road driving scenarios, packet loss can be severe since vehicles usually move fast and traffic can be crowded \cite{Multi2015Shen}, which can incur unreliable wireless communications. A framework for cooperative localization, which is robust against communication failure, is desired.

\begin{figure}[t]
\includegraphics[width=3.4in]{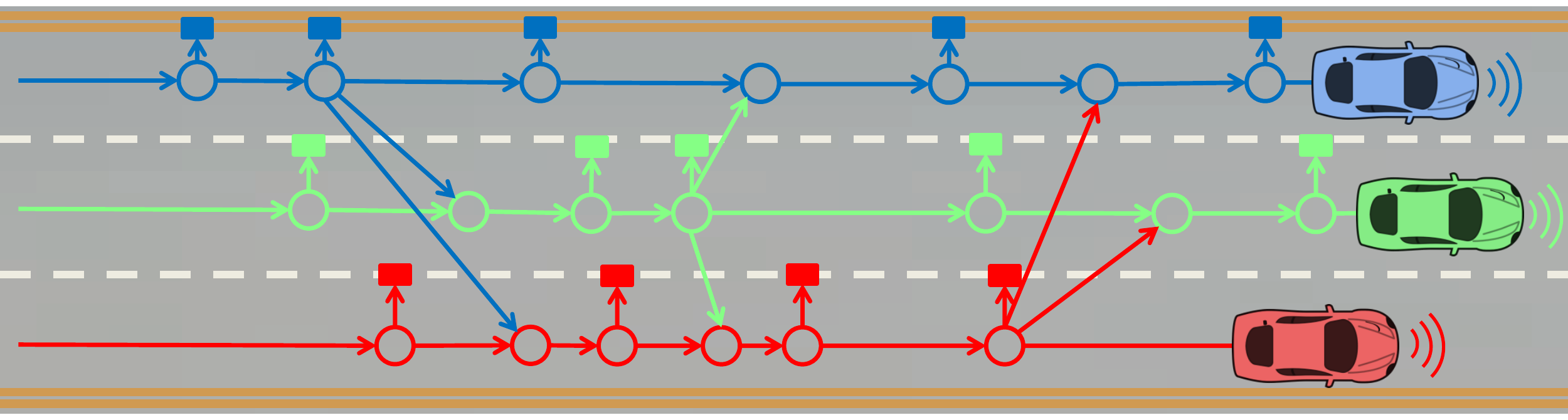}
\caption{An example of three vehicles' cooperative localization using pose graph and vehicle-to-vehicle (V2V) communication. A circle represents a vehicle's pose at a particular time, while a square denotes a map measurement, from which the pose relative to the map can be estimated. The arrow between two poses of the same vehicle represents an odometry measurement, while the arrow between two poses of difference vehicles represents a relative observation. The relative observation between vehicles correlates their pose estimation in the cooperative localization.}
\label{fig_snapshot_3vehicle_CL}
\end{figure}

Cooperative localization using pose graph can handle communication delay and out-of-sequence measurements since the delayed information or out-of-sequence measurements can be added into the pose graph as links between pose nodes \cite{Origin2014Walls}. Essentially, the pose graph can store the received information for a certain period and if the delayed information is received within this period, it can be added into the pose graph. An example of cooperative localization using pose graph is shown in Fig.\ \ref{fig_snapshot_3vehicle_CL}. 
In order to account for communication loss, Walls \textit{et al.} propose a factor composition/decomposition to recover odometry factors, so that the pose graph is always a connected component. 

To add relative observations into a pose graph, we need to know the identity of the observee vehicle, so that we know which two vehicles are associated with a particular relative observation. In other words, it is necessary to know vehicle identity so that we can match what vehicle ``sees" (by sensors) and what vehicle ``hears" (by communications).  The state-of-the-art cooperative localization approach \cite{Origin2014Walls} uses distinctive transmission signals for vehicle identification. However, for V2V communication, all the vehicles use the same spectrum and band, therefore it is challenging for vehicle identification. In this paper, we propose a general framework of cooperative localization, which is not only robust against communication delay and loss, but also solves the vehicle identification problem using data association techniques. Besides, our framework is able to accommodate more information to be fused, compared with that of \cite{Cooperative2015Walls} where the correlations between poses of different server vehicles are not utilized in the server-client scheme, as explained in Section II.

\begin{figure}[!htb]
\centering
\includegraphics[width=3.0in]{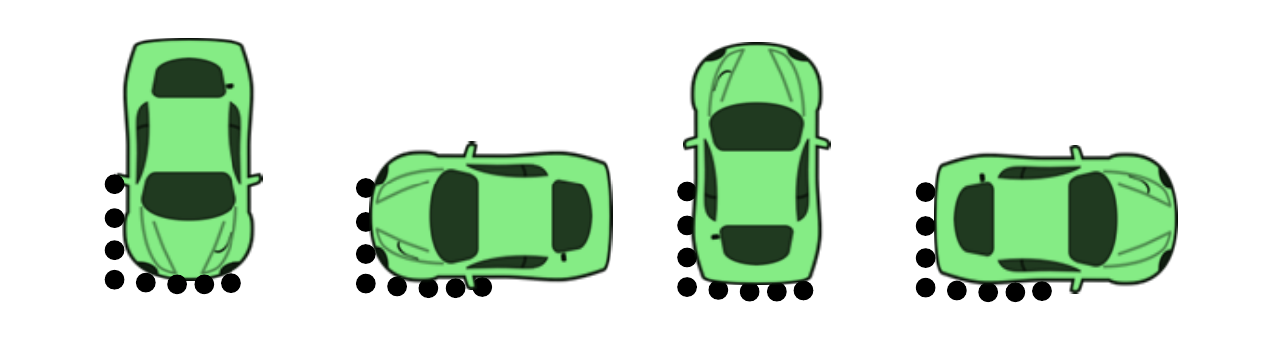}
\caption{The non-uniqueness of vehicle pose estimate using L-shape fitting. The black dots represent the LIDAR scan points. The position and orientation cannot be uniquely determined from the L-shape.}
\label{fig_non_uniqueness_of_vehicle_pose_estimate}
\end{figure}

Another challenge for cooperative localization is to obtain accurate relative observations. Even though a vehicle contour can be detected with some range sensors and an L-shape can be fitted \cite{EfficientL2015Shen}, the estimated relative pose cannot be uniquely determined and thus is ambiguous, as shown in Fig.\ \ref{fig_non_uniqueness_of_vehicle_pose_estimate}. If one adds $\pm90$ or $\pm180$ degrees to the estimated yaw angle, the L-shape would still look the same. Our proposed framework takes multiple hypotheses on the relative pose estimation into account and can solve the ambiguity before adding it to the pose graph.

The contribution of this paper can be summarized as follows:
\begin{itemize}
	\item We present a general framework for multi-vehicle cooperative localization using pose graph, which is not only robust against packet loss and communication delay but also able to accommodate more sensing measurements.
	\item Vehicle identification is formulated as a Linear Programming (LP) problem, which matches what vehicle ``sees" (by sensors) and what vehicle ``hears" (by communications).
	\item The cooperative localization results are utilized for resolving the ambiguities in vehicle relative pose estimation before adding the relative pose information into the pose graph.
\end{itemize}

The remainder of this paper is organized as follows. Section II presents the related works. Section III introduces our general framework for multi-vehicle cooperative localization. The corresponding algorithms for cooperative localization are proposed in Section IV. Section V provides the experimental results of multi-vehicle cooperative localization with V2V communication. Section VI concludes this paper.

\section{Related Works}
\subsection{Communication Constraints}
\begin{figure}[!htb]
\centering
\includegraphics[width=3.4in]{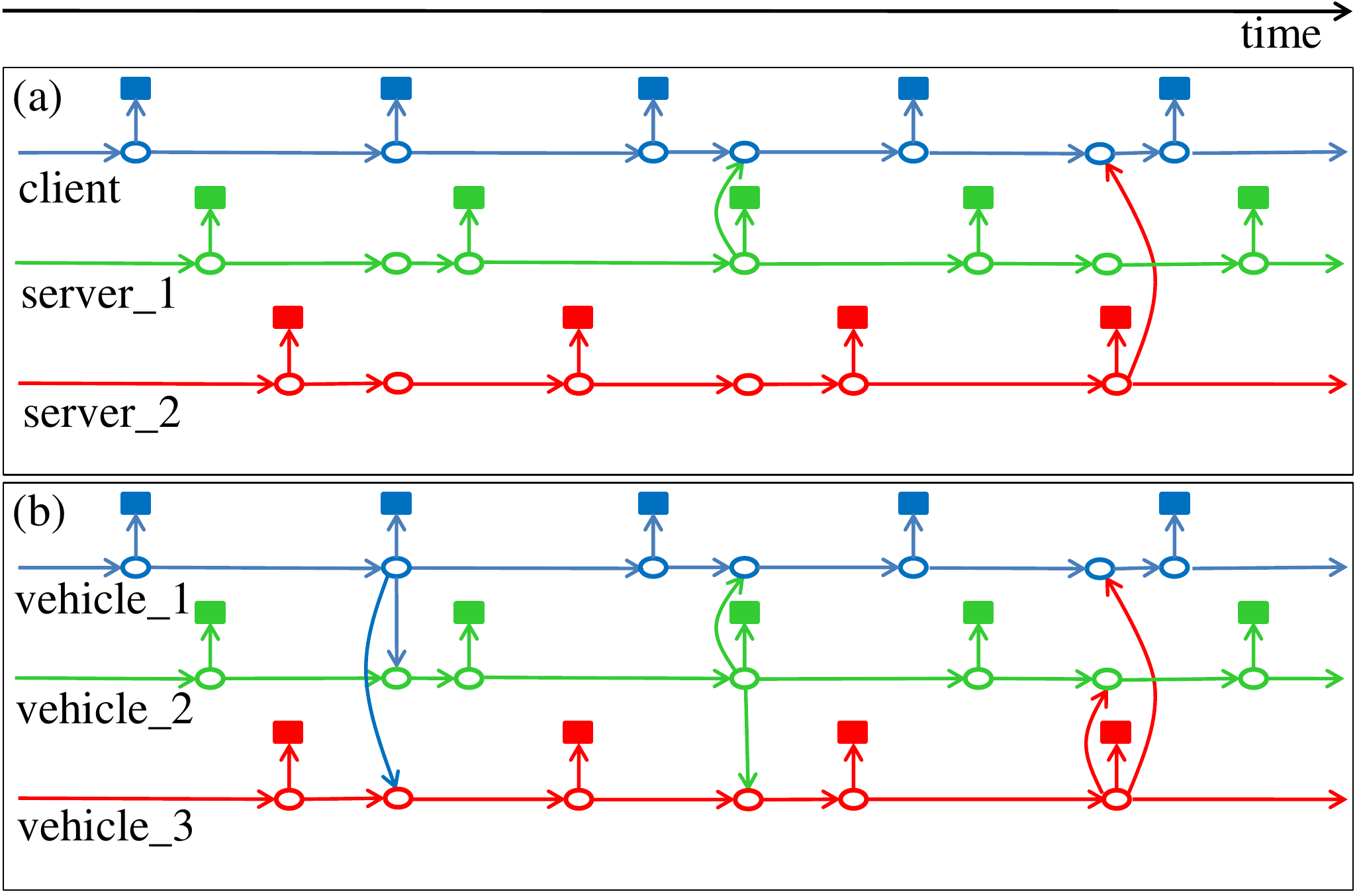}
\caption{The received measurements by the client vehicle (a) and by vehicle\_1 (b). In (a), only relative measurements observed from server vehicles are added into the pose graph \cite{Origin2014Walls}. In (b), all the relative measurements are shared with V2V communication and then added into the pose graph in our framework.}
\label{fig_subset_of_measurements}
\end{figure}

Communication constraints are mostly considered in the application of underwater robotics, where wireless communication is extremely unreliable. The origin state method is able to recover the odometry factor approximately when the packet is lost in the network where no acknowledgment is assumed \cite{Origin2014Walls}. This method is further improved with a scheme of factor composition in which origin shifting is removed \cite{Cooperative2015Walls,Decentralized2014Paull}. However, in such approaches, each vehicle receives only a subset of all the measurements since only relative measurements observing from server vehicles are added into the pose graph and the relative observations between servers are not sent to client vehicles, as shown in Fig.\ \ref{fig_subset_of_measurements}(a). Essentially, the correlation between the poses of server vehicles are not utilized for cooperative localization, which is due to the use of One-Way-Travel-Time (OWTT) for both distance measuring and communication.

The delayed-state information filter was proposed to take communication delay into account so that the delayed information can still be used in the filter \cite{delayed2009Capitan}. However, it only considers the cooperative tracking cases where the control input information of the tracked object is assumed to be inaccessible. In this paper, we are proposing a framework for cooperative localization, which not only considers packet loss and communication delay, but also accommodates all the shared sensing information, as shown in Fig.\ \ref{fig_subset_of_measurements}(b).
\subsection{Optimal filtering}
The optimal filtering with all the shared information is one of the key steps for cooperative localization. Bahr \textit{et al.}~\cite{Consistent2009Bahr} proposed a consistent cooperative localization approach to bookkeep the origins of measurements and thereby prevent the use any of the measurements more than once. More common approaches \cite{Cooperative2013Li,Decentralized2010Leung} utilize covariance intersection (CI) techniques to avoid overconfident estimate. With CI techniques, it is unnecessary for each vehicle to know the communication topology and cooperative localization can be performed in a decentralized manner. However, the correlation of the measurements is not fully exploited by CI techniques and thus the estimate is suboptimal. Zhang \textit{et al.}~\cite{MultiCL2013Zhang} proposed to utilize Probability Hypothesis Density (PHD) filter for cooperative localization. However, the computational complexity makes it less scalable. Non-linear least square optimization technique is utilized to obtain the optimal estimate in \cite{Origin2014Walls,Decentralized2014Paull,Cooperative2015Walls}. In this paper, we also formulate the cooperative localization as a non-linear least square optimization problem. The graphical model, shown in Fig. \ref{fig_snapshot_3vehicle_CL}, essentially explains the dependencies of all the measurements. The sparsity of the graph allow us to perform inference efficiently. Any available efficient non-linear least square optimization solver, such as g2o \cite{g2o2011Kummerle} and iSAM \cite{iSAM2008Kaess}, can be used to obtain the optimal estimate.

\subsection{Vehicle Identification}
It is difficult and challenging to identify who sent a message that arrived at ego vehicle because V2V communication devices are typically omni-directional and use the same spectrum and band \cite{Cooperative2013Kim}. On the other hand, it is important to identify the IDs of the detected vehicles so that we know which two vehicles are associated with a relative observation. The principle way of vehicle identification is matching some common information between what the vehicle recognizes by perception and what the vehicle receives by communication. For instance, QR codes are placed on a vehicle for other vehicles recognizing its ID with a camera \cite{olson2013exploration}. Besides, license plate number can be recognized as vehicle ID using cameras \cite{chang2004automatic,hongliang2004hybrid,anagnostopoulos2006license}. These vision methods can suffer from bad lighting conditions, motion blur, and wrong recognitions. Kim \textit{et al.} propose to match the detected speed profiles with the received ones for vehicle identification \cite{Cooperative2013Kim}. Nonetheless, the speed profiles may not be distinctive since vehicles may follow the same speed patterns on the same road. In this paper, we use vehicle pose on a global map as the common information for vehicle identification. Essentially, we match the vehicle pose detected by a LIDAR sensor with the received vehicle pose from V2V communication. For identifying multiple vehicles simultaneously, it is formulated as a linear assignment problem, which can be solved efficiently using linear programming (LP) solvers.

\section{General Framework}
\begin{figure}[!htb]
	\centering
	\includegraphics[width=3.5in]{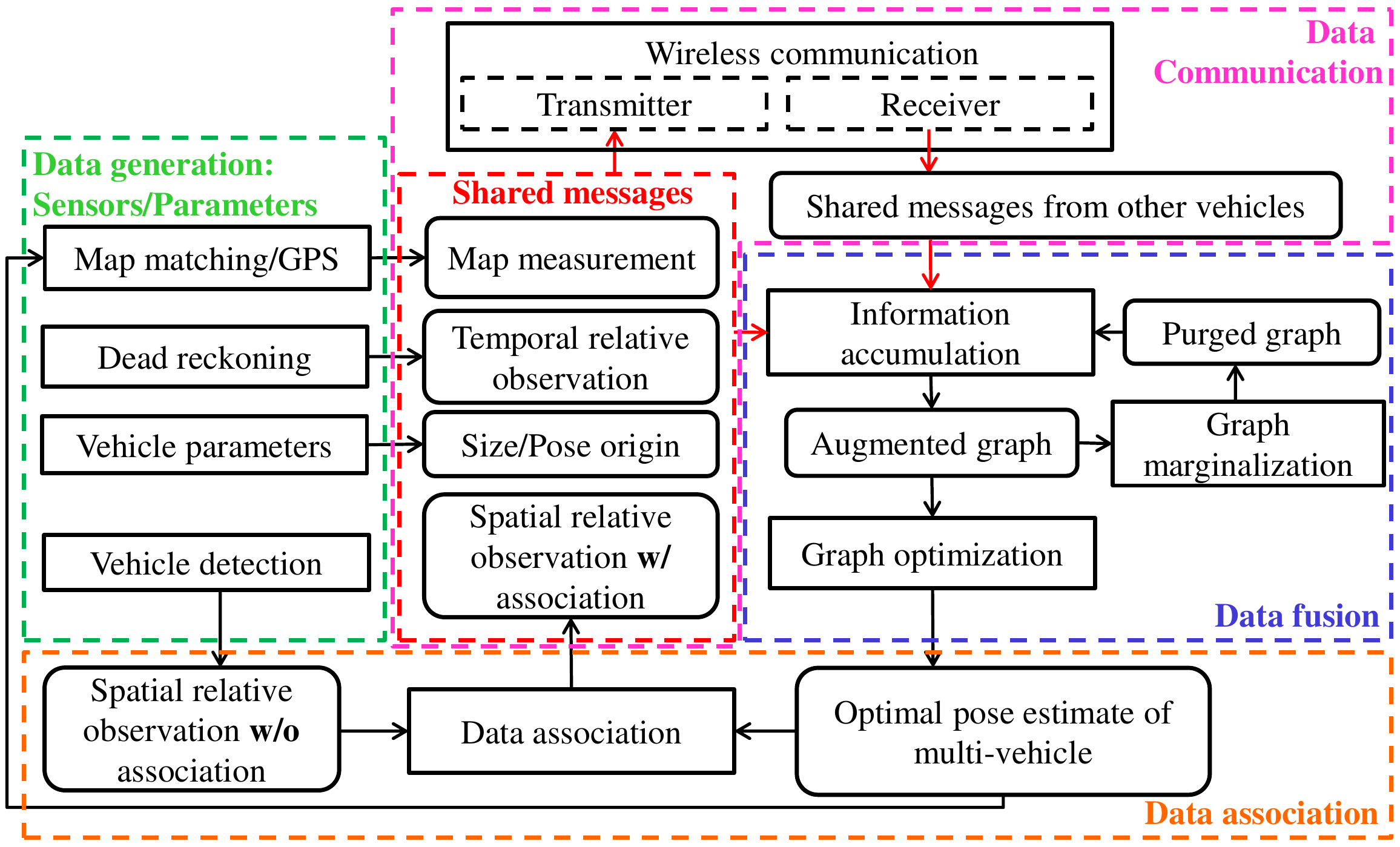}
	\caption{The framework of cooperative localization using pose graph.}
	\label{fig_system_architecture_for_cooperative_localization}
\end{figure}

Fig.\ \ref{fig_system_architecture_for_cooperative_localization} shows the proposed framework for multi-vehicle cooperative localization. The framework consists of four main modules: data generation, data communication, data fusion, and data association.

\subsection{Data Generation}
The shared data are mostly generated by the on-board sensors. In cooperative localization, the global pose can be estimated with a Global Positioning System (GPS) sensor or by matching the LIDAR scan with a global map \cite{chong2013synthetic}. With a GPS sensor or a map matching technique, the pose with respect to a global coordinate frame can be estimated and thereby the map measurements can be generated. The dead reckoning system, consisting of direct accumulated measurements from encoders and an Inertial Measurement Unit (IMU), can estimate the transform between two poses of a single vehicle at different timestamps, i.e., the temporal relative observation. With a LIDAR sensor, the nearby vehicles can be detected and the relative pose can be estimated. Nonetheless, the unique identity of the observed vehicle is still unknown from solely the vehicle detection. The vehicle identification is solved by the data association module, where the spatial relative observation is correlated to the best matching vehicle association.  Some important vehicle parameters, such as the size and the anchor point of the vehicle, are shared to facilitate the inference of the vehicle pose from the vehicle contour information.

\subsection{Data Communication}
The shared data are exchanged via wireless communication, such as 3G \cite{Cooperative2013Kim}, 4G \cite{Tele2014Shen}, WIFI \cite{Scalable2016Shen}, or V2V IEEE 802.11p \cite{Multi2015Shen}. There are four types of information to be shared: map measurement, temporal relative observation, vehicle geometrical parameters, and spatial relative observation with association. These data are broadcast to all the nearby vehicles through the wireless communication module.

\subsection{Data Fusion}
After receiving the shared information from nearby vehicles, the data fusion module first accumulates them into a pose graph and augments the historical pose graph with the new data. A pose graph optimization algorithm, such as iSAM, is carried out to obtain optimal pose estimation. The graph size would keep on increasing with the arrival of new information. In order to bound the graph size, the graph should be purged by removing the old pose nodes. The generic node removal method can be adopted to marginalize the pose graph \cite{carlevaris2014generic}.

\subsection{Data Association}
With the optimal poses extracted from the data fusion module, the poses are associated with the L-shapes detected by the LIDAR sensor. The data association module essentially corresponds each detected L-shape to a unique vehicle. With the established correspondence, the spatial relative observation between each vehicle is then known. The corresponded spatial relative observation is then broadcast to be added into the pose graph.

It should be highlighted that with a more accurate localization result, the data association is easier to be determined since the correspondence is more discriminant when localization uncertainty is low. In turn, a correct data association can help to add the inter-vehicle relative observation into the graph for further improving localization accuracy. Meanwhile, a more accurate localization result also helps the map matching since the initial guess is already accurate and certain to some extent. A better map matching also would help adding the correct map measurements into the pose graph, which in turn improves the localization accuracy.

\section{Proposed Algorithms}
In this section, we propose the algorithms for two core modules of cooperative localization, data fusion and data association. Before that, we introduce some notations.
\subsection{Notations}
In this paper, we only consider 2D vehicle pose $(x,y,\theta)^T$, where $(x,y)^T$ denotes the 2D position and $\theta$ denotes the orientation. $^{i}S(t)=(^ix(t),^iy(t),^i\theta(t))^T$ denotes the global pose of vehicle $i$ at time $t$. The temporal relative observation is essentially the relative pose of a single vehicle between two different timestamps, which can be described as follows,
\begin{equation}
\label{eqn_cl_temporal_relative_observation}
\leftidx{^i}R(t,t') = \leftidx{^i}S(t') \ominus  \leftidx{^i}S(t),
\end{equation}
where $\leftidx{^i}R(t,t')$ represents the pose of vehicle $i$ at time $t'$ relative to the pose of the same vehicle at time $t$, and $\ominus$ operator represents the computation of the relative transformation of two poses \cite{castellanos2006map}. The spatial relative observation is essentially the relative pose between two vehicles at a specific time $t$, which can be described as follows,
\begin{equation}
\label{eqn_cl_spatial_relative_observation}
\leftidx{^{ij}}R(t) = \leftidx{^j}S(t) \ominus \leftidx{^i}S(t),
\end{equation}
where  $\leftidx{^{ij}}R(t)$ denotes the pose of vehicle $j$ relative to the pose of vehicle $i$ at time $t$.

In order to account for the uncertainty of pose estimation, these three types of measurements are assumed to be characterized by normal distributions. Specifically, we assume $^{i}S(t) \sim \mathcal{N}(^{i}\mu(t),\leftidx{^i}\varSigma(t))$, $\leftidx{^i}R(t,t')  \sim \mathcal{N}(^{i}\mu(t,t'),\leftidx{^i}\varSigma(t,t'))$, $\leftidx{^{ij}}R(t) \sim \mathcal{N}(\leftidx{^{ij}}\mu(t), \leftidx{^{ij}}\varSigma(t))$, where $^{i}\mu(t)$, $^{i}\mu(t,t')$, and $\leftidx{^{ij}}\mu(t)$ represent the mean, while $\leftidx{^i}\varSigma(t)$, $\leftidx{^i}\varSigma(t,t')$, and $\leftidx{^{ij}}\varSigma(t)$ represent the covariance, of $^{i}S(t)$, $\leftidx{^i}R(t,t')$, and $\leftidx{^{ij}}R(t)$, respectively. These three types of information compose the basic elements of a pose graph, which is shown in Fig.\ 5. Combining these data together, we can have a pose graph, as shown in Fig.\ \ref{fig_snapshot_3vehicle_CL}. 

\begin{figure}[!thb]
	\label{fig_cl_three_types_information}
	\begin{center}
		\subfigure[map measurement $^{i}S(t)$.]{
			\label{fig:subfig:map_measurement}
			\includegraphics[height=0.9in]{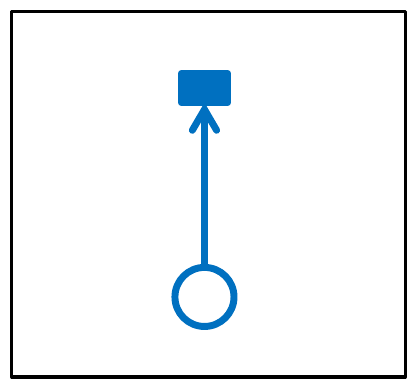}}
		\hspace{0.03in}
		\subfigure[temporal relative observation ${^i}R(t,t')$.]{
			\label{fig:subfig:temporal_relative_observation}
			\includegraphics[height=0.9in]{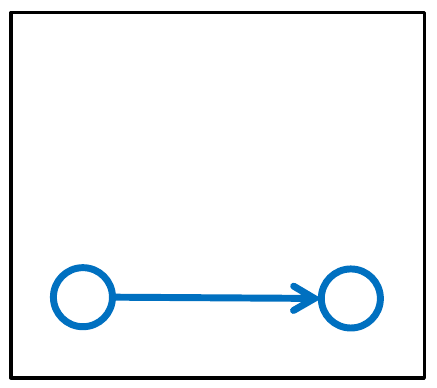}}
		\hspace{0.03in}
		\subfigure[spatial relative observation $^{ij}R(t)$.]{
			\label{fig:subfig:spatial_relative_observation}
			\includegraphics[height=0.9in]{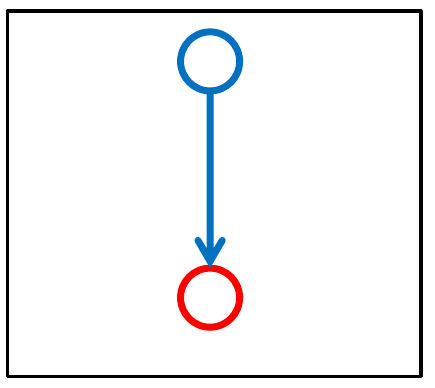}}
		\hspace{0.03in}
		\caption{Three types of information in a pose graph.}
	\end{center}
\end{figure}

\subsection{Shared information}
Localization is essentially the process to estimate the vehicle pose at a specific time. The timestamp is thus necessary for knowing what time an estimated pose corresponds to. We assures that all the vehicle clocks are synchronized with a global server. In the map measurements, the packet contains the following elements: vehicle ID, measurement timestamp, the mean of the pose, and the covariance of the pose. More formally, the measurement packet contains $\{i,t,\leftidx{^i}\mu(t),\leftidx{^i}\varSigma(t)\}$. The map measurement can be generated by a scan matching algorithm \cite{chong2013synthetic} or a GPS sensor. 

A simple way of sharing the temporal relative observation $\mathcal{N}(^{i}\mu(t,t'),\leftidx{^i}\varSigma(t,t'))$ is just broadcasting it. However, if the communication fails, the odometry chain will break since the link between the two poses at $t$ and $t'$ is missing. The whole pose graph is in risk of breaking into multiple disconnected components, if no acknowledgment and re-broadcasting is carried out. Walls \textit{et al.} proposed to broadcast $\mathcal{N}(^{i}\mu(0,t),\leftidx{^i}\varSigma(0,t))$ instead. With such a scheme, even if the information $\mathcal{N}(^{i}\mu(0,t_2),\leftidx{^i}\varSigma(0,t_2))$ is lost, the relative observation between two poses at $t_1$ and $t_3$ can still be recovered with a factor decomposition \cite{Cooperative2015Walls} if the factors at $t_1$ and $t_3$ are received. In this paper, we also adopt this scheme so that our framework is robust against communication failure. Therefore, in the spatial relative observation packet, we broadcast $\{i,t,\leftidx{^i}\mu(0,t),\leftidx{^i}\varSigma(0,t)\}$. The temporal relative observation can be obtained from the odometry sensors, such as an IMU and encoders.

The spatial relative observation involves two different vehicles, and thus the two vehicles' IDs need to be broadcast. Usually, a spatial relative pose is observed instantly with a LIDAR sensor and thus a single timestamp is sufficient to describe the related poses. In the spatial relative observation, we broadcast $\{i,j,t,\leftidx{^{ij}}\mu(t),\leftidx{^{ij}}\varSigma(t)\}$, where $i$ represents the observer vehicle ID and $j$ represents the observee vehicle ID. The spatial relative observation is obtained through the inference on the relative pose from the L-shape \cite{EfficientL2015Shen}. The vehicle ID is assigned by a data association algorithm, which is described in Section \ref{data_assoc_subsection}.

\subsection{Localization Algorithm}
The proposed cooperative localization algorithm essentially fuses all the shared information to a single pose graph. The pose graph based algorithm is described in Algorithm \ref{algo_coop_localiztion_pose_graph}. The pose graph $G$ is empty initially. After receiving map measurements and spatial relative observations, they are added to the graph $G$ as edges by the function \texttt{AddMapMeas} and \texttt{AddSpatialRelObs} respectively. For temporal relative observation, the pose at time $t$ relative to the initial pose at time $0$ is transmitted in order to be robust against communication failure \cite{Cooperative2015Walls}. Once the temporal relative observation is received, it will be decomposed so that the pose is relative to the latest pose at time $t'$ in the odometry chain using the factor decomposition function \texttt{FacDecomp}. The decomposition method is proposed in \cite{Cooperative2015Walls}. After decomposition, the temporal relative observations are then added into the graph by the function \texttt{AddTemporalRelObs}.

\begingroup
\SetInd{0.01em}{0.01em}
\begin{algorithm}[!thb]
\SetAlgoLined
\SetKwFunction{AddMapMeas}{AddMapMeas}
\SetKwFunction{AddTemporalRelObs}{AddTemporalRelObs}
\SetKwFunction{AddSpatialRelObs}{AddSpatialRelObs}
\SetKwFunction{GraphOptimization}{GraphOptimization}
\SetKwFunction{FacDecomp}{FacDecomp}
\SetKwFunction{RemoveOldNodes}{RemoveOldNodes}
\SetKwInOut{Input}{input}\SetKwInOut{Output}{output}

\Indm\Indmm
\Input{$G,i,j,t,\leftidx{^i}\mu(t),\leftidx{^i}\varSigma(t),\leftidx{^i}\mu(0,t),\leftidx{^i}\varSigma(0,t),\leftidx{^{ij}}\mu(t),\leftidx{^{ij}}\varSigma(t)$}
\Output{$\{ \leftidx{^i}{\hat{S}}(t) | \forall i \in [1,n]\}$, $G$}
\BlankLine
// pose graph construction \\
$G \leftarrow$ \AddMapMeas{$G,i,t,\leftidx{^i}\mu(t),\leftidx{^i}\varSigma(t)$}

\Indm$(\leftidx{^i}\mu(t',t),\leftidx{^i}\varSigma(t',t)) \leftarrow$ \FacDecomp{$G,\leftidx{^i}\mu(0,t),\leftidx{^i}\varSigma(0,t)$}

$G \leftarrow$ \AddTemporalRelObs{$G,i,t,\leftidx{^i}\mu(t',t),\leftidx{^i}\varSigma(t',t)$}

$G \leftarrow$ \AddSpatialRelObs{$G,i,j,t,\leftidx{^{ij}}\mu(t),\leftidx{^{ij}}\varSigma(t)$}

// pose graph optimization

$\{ \leftidx{^i}{\hat{S}}(t) | \forall i \in [1,n]\} \leftarrow$ \GraphOptimization{$G$}

// pose graph marginalization

$G \leftarrow $  \RemoveOldNodes{$G$}

\KwRet{$\{ \leftidx{^i}{\hat{S}}(t) | \forall i \in [1,n]\},G$}

\AlgoDisplayBlockMarkers

\caption{Cooperative Localization Using Pose Graph}
\label{algo_coop_localiztion_pose_graph}
\end{algorithm}
\endgroup

With the pose graph being augmented with new measurements, a pose graph optimization algorithm, such as iSAM, is carried out to extract the latest estimated poses $\{ \leftidx{^i}{\hat{S}}(t) | \forall i \in [1,n]\}$ for all the vehicles, indexed from $1$ to $n$. With the accumulation of the measurements, the size of the pose graph would grow. The size growing would slow the graph optimization and eventually use up all the computer memory. In order to maintain a moderate size of the pose graph, the old nodes should be removed without disregarding their significance. A generic node removal algorithm \cite{carlevaris2014generic} is proposed, which ensures that the optimization result would be the same when the nodes are removed. It should be highlighted that the size of the graph is in correlation with the time window of the graph, which determines the ability of handling the delayed measurements. If the timestamp of a measurement is out of the time window, it will not be able to be inserted in the pose graph since the old nodes are already marginalized out. After node removal, the pose graph $G$ is updated and stored.

Algorithm \ref{algo_coop_localiztion_pose_graph} essentially is the back-end for cooperative localization using pose graph. The front-end would be obtaining the three types of measurements. Map measurement and temporal relative observation can be obtained with traditional localization methods \cite{chong2013synthetic,Scalable2016Shen} since they are basic elements for independent localization. The essence of cooperative localization is to exploit the spatial relative observation to correlate the poses between vehicles. The acquisition of the spatial relative observation for each vehicle is described in Section \ref{data_assoc_subsection}.

\subsection{Data Association}\label{data_assoc_subsection}
The method for estimating relative pose from an L-shape is introduced in \cite{EfficientL2015Shen}, where the ambiguities in both the corner point and the orientation is considered. In order to add the relative pose information into the graph, we need to choose one of the hypothetical relative poses. Even though the best L-shape, in terms of the shape fitting error, can be determined, there are still four possible poses due to the ambiguity in the orientation, as shown in Fig.\ \ref{fig_non_uniqueness_of_vehicle_pose_estimate}. Moreover, it is unknown which vehicle corresponds to the relative pose considered since the vehicle identity cannot be detected solely from the L-shape. Data association is critical in determining the correspondence between the L-shape and the vehicle ID. 

Since the data association algorithms are the same on each vehicle, we only consider the data associations between the $l$-th vehicle's observations and the other vehicles. Assuming that there are $m$ detected L-shapes $\{\mathcal{H}_1, \cdots, \mathcal{H}_m \}$ from vehicle $l$ and $n$ vehicles $\{\leftidx{^1}{\hat{S}}(t), \cdots,\leftidx{^n}{\hat{S}}(t) \}$, the data association is essentially to determine the correspondence $\{k_1,\cdots,k_n\}$ that pairs each estimated vehicle pose $\leftidx{^i}{\hat{S}}(t)$  with an L-shape $\mathcal{H}_{k_i}$, where $k_i \in [1,m], i \in [1,n]$. The vehicle pose $\leftidx{^i}{\hat{S}}(t)$ is estimated from the cooperative localization algorithm in Algorithm \ref{algo_coop_localiztion_pose_graph}. For each L-shape $\mathcal{H}_k$, $k \in [1,m]$, there are multiple hypotheses on estimated relative poses of the vehicle corner, $h_{k}(v) = (x_{k}(v),y_{k}(v),\varTheta_k(v))^T$, where $v\in[1,d_k]$ is the index, $d_k$ is the number of hypothetical corner points in L-shape $\mathcal{H}_k$, $(x_{k}(v),y_{k}(v))^T$ is the vehicle corner point, and $\varTheta_k(v) = \{\theta_{0k}(v),\theta_{1k}(v),\theta_{2k}(v),\theta_{3k}(v)\}$ is the set of four possible vehicle orientations. Broadcast through wireless communication, the vehicle size and origin pose can be retrieved, and thereby the transformations between the four corner points and the vehicle origin can be derived. We denote the four transformations as $\leftidx{^i}Q=\{q_0(i),q_1(i),q_2(i),q_3(i)\}$ of vehicle $i$, where $q_p (i)= (\tilde{x}_{p}(i),\tilde{y}_{p}(i),\tilde{\theta}_{p}(i))^T $ denotes the transform from the corner point pose to the vehicle origin pose, and $p \in [0,3]$ represents the index of four possible vehicle corners. With the hypothetical pose $h_{k}(v)$ and the transform $\leftidx{^i}Q$, we can infer the possible vehicle poses of the origin $\leftidx{^i}{\tilde{S}}(k,v,p)$ as follows,
\begin{equation}
\label{eqn_inferring_vehicle_pose_from_corner_point}
\leftidx{^i}{\tilde{S}}(k,v,p) = (x_{k}(v),y_{k}(v),\theta_{pk}(v))^T \oplus q_p(i), p\in[0,3],
\end{equation}
where the operator $\oplus$ denotes the transformation composition \cite{castellanos2006map}, $k\in[1,m],v\in[1,d_{k}]$. The association cost $\Phi_{ik}$ of the pairing $(i,k)$ is defined as the minimal ``distance" between the pose estimated from the L-shape $\mathcal{H}_k$ and the latest pose $\leftidx{^i}{\hat{S}}(t)$ estimated from the pose graph $G$, which is computed as follows,
\begin{equation}
\label{eqn_data_association_cost}
\Phi_{ik} = \min_{v,p} \texttt{dist}(\leftidx{^i}{\tilde{S}}(k,v,p), \leftidx{^i}{\hat{S}}(t)),
\end{equation}
where $v \in [1,d_{k}], p \in [0,3], i\in[1,n],k\in[1,m]$, \texttt{dist} is the distance function between two poses. The distance function can be computed as follows,
\begin{equation}
\begin{aligned}
\texttt{dist}((x_1,y_1,\theta_1)^T,(x_2,y_2,\theta_2)^T) & =  \lVert(x_1-x_2,y_1-y_2)\rVert_{2}^{2} \\&+ W * g(\theta_1 - \theta_2),
\end{aligned}
\end{equation}
where $\lVert \cdot \rVert_2$ is the Euclidean norm, $W$ is the weight for the squared angular distance and $g$ computes the square of the normalized yaw angle difference as follows,
\begin{equation}
g(\Delta \theta) = \left(\texttt{atan2}\left(\texttt{sin}(\Delta \theta),\texttt{cos}(\Delta \theta)\right) \right)^2.
\end{equation}

It is possible that vehicle $i$ is not detected and thus not associated with any L-shape. In order to take this possibility into account, we define $n$ null observations $\{\mathcal{O}_1, \cdots, \mathcal{O}_n\}$ to augment the observation set as $\{ \mathcal{H}_1,\cdots, \mathcal{H}_m, \mathcal{O}_1, \cdots, \mathcal{O}_n \}$. The cost $\Phi_{ik}$ for associating vehicle $i$ with the $k-$th observation is defined in \eqref{eqn_data_association_cost} when $k \in [1,m]$, while $\Phi_{ik}$, $k\in[m+1,m+n]$, represents the association cost for associating vehicle $i$ with $(k-m)$-th null observation, i.e., the $k$-th observation of the augmented observation set, which is defined as follows,
\begin{equation}
\label{eqn_data_association_null_observation_cost}
\Phi_{ik} = \begin{cases}
\varUpsilon, & i = k-m,\\
\infty, & \text{otherwise},
\end{cases}
\end{equation}
where $k \in [m+1,m+n]$, $i\in[1,n]$, $\varUpsilon$ is a constant cost value. Putting the association cost $\Phi_{ik}$ in \eqref{eqn_data_association_cost} and \eqref{eqn_data_association_null_observation_cost} into the assignment matrix $\mathbf{\Phi} = [\Phi_{ik}] \in \mathbb{R}^{n \times (m+n)}$, the data association problem can be formulated as a linear assignment problem \cite{martello1987linear} as follows,
\begin{equation*}
\begin{aligned}
& \underset{}{\text{minimize}}
& & \sum_{i=1}^{n} \sum_{k=1}^{m+n} z_{ik} \Phi_{ik} \\
& \text{subject to}
& & z_{ik} \in \{0,1\}, \sum_{i=1}^{n} z_{ik}  \leq 1, \sum_{k=1}^{m+n} z_{ik}  = 1,
\end{aligned}
\end{equation*}
where $z_{ik}$ is an indicator variable, $z_{ik}=1$ represents that vehicle $i$ is associated with observation $k$ and $z_{ik}=0$ represents otherwise. The problem is an integer program \cite{bertsimas1997introduction}, but due to the special structure it can be solved efficiently using linear programming (LP) and the obtained optimal solution $\mathbf{z}^* = [z_{ik}^*]$ is guaranteed to be in $\{0,1\}^{n\times(m+n)}$. The correspondence $\{k_1,\cdots, k_n\}$ can then be determined as follows,
\begin{equation}
\label{eqn_extracting_correspondence_index}
k_i = \arg \max_{k} z^*_{ik},
\end{equation}
where $k_i$ represents the index of the observation. If $k_i \in [1,m]$, it means that vehicle $i$ is corresponding to the L-shape $\mathcal{H}_{k_i}$. If $k_i \in [m+1,m+n]$, it means that vehicle $i$ is corresponding to a null observation, i.e., is not observed. Since we assume that all the L-shapes in $\{\mathcal{H}_1,\cdots,\mathcal{H}_m\}$ are detected from vehicle $l$, the data association essentially determines that vehicle $l$ observes vehicle $i$ as an L-shape $\mathcal{H}_{k_i}$ if $k_i \in [1,m]$.  In order to have a complete spatial relative observation $\{l,i,t,\leftidx{^{li}}\mu(t),\leftidx{^{li}}\varSigma(t)\}$, we still need to compute $\leftidx{^{li}}\mu(t)$ and $\leftidx{^{li}}\varSigma(t)$ from both $\mathcal{H}_{k_i}$ and the pose $\leftidx{^i}{\hat{S}}(t)$ by the method in Section \ref{relative_pose_estimation_subsection}.
\subsection{Relative Pose Estimation}\label{relative_pose_estimation_subsection}
Assuming that the L-shape $\mathcal{H}_{k}$ observed from vehicle $l$ is associated with vehicle $i$, we need to select the best hypothetical pose from $\{\leftidx{^i}{\tilde{S}}(k,v,p)| v \in[1,d_{k}], p \in[0,3] \}$ for estimating $\{\leftidx{^{li}}\mu(t),\leftidx{^{li}}\varSigma(t)\}$. The criterion for determining the best hypothesis should depend on the L-shape fitting error $\lambda_{v}$. However, for the same fitting error $\lambda_{v}$, there are still four possible orientations. In order to handle the ambiguities in both the corner point index and orientation, we define the criterion function $f$ as follows,
\begin{equation}
\label{eqn_criterion_for_selecting_hypothesis}
\begin{aligned}
f(\lambda_v,&\leftidx{^i}{\tilde{S}}(k,v,p), \leftidx{^i}{\hat{S}}(t)) = w_1 * \lambda_v \\ & + w_2 * g\left( e_3^T \left(  \leftidx{^i}{\tilde{S}}\left(k,v,p\right) -   \leftidx{^i}{\hat{S}}\left(t\right)\right)\right),
\end{aligned}
\end{equation}
where $\lambda_v$ is the L-shape fitting error of $\leftidx{^i}{\tilde{S}}(k,v,p)$ as computed in \cite{EfficientL2015Shen}, $e_3=(0,0,1)^T$, $w_1 > 0 $ is the weight for the L-shape fitting error, and $w_2>0$ is the weight for the squared orientation difference between $\leftidx{^i}{\tilde{S}}(k,v,p)$ and $\leftidx{^i}{\hat{S}}\left(t\right)$, respectively. Essentially, the orientation difference is more significantly considered when the fitting errors are the same or close. With the incorporation of the pose information $\leftidx{^i}{\hat{S}}\left(t\right)$ from cooperative localization, the ambiguities in $v$ and $p$ can be resolved much more easily as follows,
\begin{equation}
\label{eqn_best_pose_determined_from_cl}
(v^*,p^*) = \arg \min_{v,p} f(\lambda_v,\leftidx{^i}{\tilde{S}}(k,v,p),\leftidx{^i}{\hat{S}}(t)),
\end{equation}
where $v^*$ represents the optimal corner point index, $p^*$ represents the optimal index for vehicle orientation. The problem \eqref{eqn_best_pose_determined_from_cl} can be solved very efficiently by iterating through all possible $v$ and $p$. The mean of the relative pose can then be computed as follows,
\begin{equation}
\label{eqn_relative_pose_mean}
\leftidx{^{li}}{\mu(t)} = \leftidx{^i}{\tilde{S}}(k,v^*,p^*) \ominus \leftidx{^l}{\hat{S}}(t) .
\end{equation}

The covariance of the relative pose can be computed in the following way: 1) uniformly sampling $N$ poses $\leftidx{^{lj}}{\tilde{R}}(t)$ around $\leftidx{^{li}}{\mu(t)}$ in the interval $[\leftidx{^{li}}{\mu(t)}-\Delta,\leftidx{^{li}}{\mu(t)} + \Delta ]$, where $j\in[1,N]$ and $\Delta$ is a constant parameter; 2) computing the sum of the squared distances from the LIDAR scan points to the corresponding lines $\lambda_j$; 3) computing the probability $ \Pr[\leftidx{^{li}}{R}(t) = \leftidx{^{lj}}{\tilde{R}}(t)]$ for each sampled pose as in \cite{EfficientL2015Shen}; 4) computing the covariance  $\leftidx{^{li}}\varSigma(t)$  as follows,
\begin{equation}
\label{eqn_relative_pose_covariance}
\begin{aligned}
\leftidx{^{li}}{\varSigma(t)} =  \sum_{j=1}^{N} \Big( & \Pr[\leftidx{^{li}}{R}(t) =  \leftidx{^{lj}}{\tilde{R}}(t)] * \\ & (\leftidx{^{lj}}{\tilde{R}}(t) -  \leftidx{^{li}}{\mu(t)})( \leftidx{^{lj}}{\tilde{R}}(t) - \leftidx{^{li}}{\mu(t)})^T \Big).
\end{aligned}
\end{equation}

In this section, the back-end algorithm of the cooperative localization is introduced in Algorithm \ref{algo_coop_localiztion_pose_graph}, which fuses all the received information to estimate the poses. The estimated poses are associated with the L-shapes, which are the relative observations from the LIDAR sensor. With the data association algorithm, the vehicle identity problem can be solved. The ambiguities in the corner point index and vehicle orientations are resolved by incorporating the estimated global pose information from cooperative localization. The determined spatial relative poses can then be added to the pose graph for further improving pose estimation. 
\section{Experimental results}
In order to show that the proposed general framework of cooperative localization can be applied to multiple vehicles and improve the localization accuracy, we simulate six autonomous vehicles moving on a straight road and on a curvy one. We also perform real experiments using three vehicles on an urban road, where V2V communication is adopted.
\subsection{Simulation}
\subsubsection{Spatial Relative Pose Estimation}
\begin{figure}[!thb]
\centering
\includegraphics[width=3.5in]{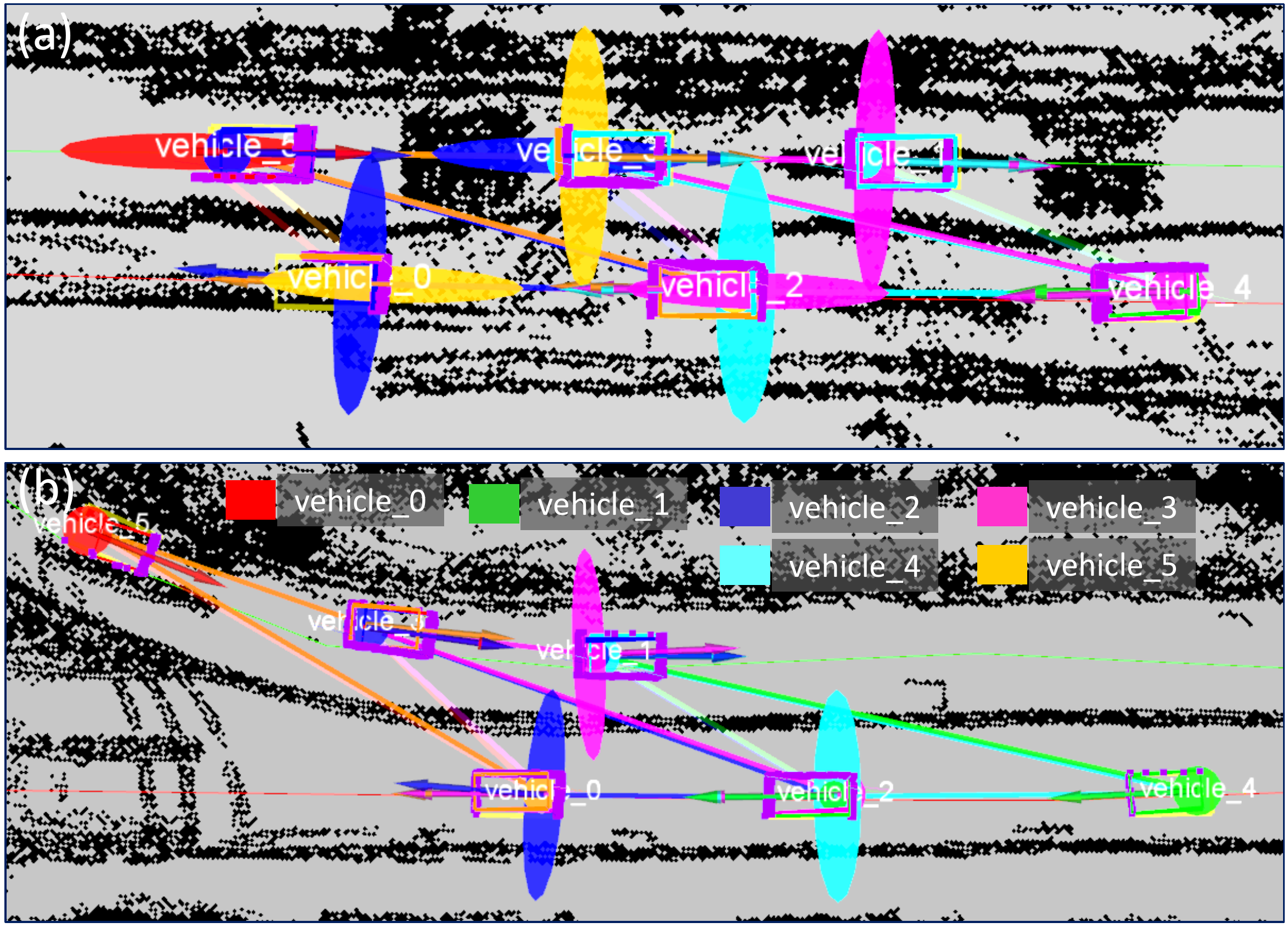}
\caption{A snapshot of spatial relative pose estimation when six vehicles move on a straight road is shown in (a), and another snapshot when vehicles move on a curvy road is shown in (b). The color label of each vehicle's detection result is shown at the top of (b). The ellipses and arrows of one color represent the relative position uncertainty and the relative orientation, estimated by the corresponding vehicle with the same color label. Each thick line of one color represents that there is a spatial relative pose estimation between the corresponding vehicle of the same color and a target vehicle. The vehicle name is printed near its contour box in (a) and (b). \text{``vehicle\_1"}, \text{``vehicle\_3"}, and \text{``vehicle\_5"} are tasked to move to the right along the green path, while \text{``vehicle\_0"}, \text{``vehicle\_2"}, and \text{``vehicle\_4"} are tasked to move to the left along the red path. The purple dots are the LIDAR scan points. }
\label{fig_multi_vehicle_relative_pose_estimation} 
\end{figure}

Each vehicle is equipped with a LIDAR sensor, which can scan around the environment with a field-of-view (FOV) of 180 degrees and is installed in the front of the vehicle. The LIDAR scans are segmented into clusters with the Adaptive Breakpoint Detector (ABD) segmentation algorithm  \cite{borges2004line, premebida2005segmentation}. The L-shape fitting algorithm \cite{EfficientL2015Shen} is carried out to generate multiple hypotheses on the relative poses. The estimated L-shapes are associated with the ID of nearby vehicles using the proposed data association algorithm, within which the linear program is solved by the GNU Linear Programming Kit (GLPK) \cite{makhorin2001gnu}. The spatial relative pose is then estimated by the proposed algorithm. The estimated spatial relative poses at two instants are shown in Fig.\ \ref{fig_multi_vehicle_relative_pose_estimation}. A snapshot of spatial relative pose estimation when six vehicles move on a straight road is shown in \ref{fig_multi_vehicle_relative_pose_estimation}(a), and another snapshot when vehicles move on a curvy road is shown in Fig.\ \ref{fig_multi_vehicle_relative_pose_estimation}(b). 

A video showing the spatial relative pose estimation while six simulated vehicles move on the road can be found at \href{https://youtu.be/lLwSusjfsbs}{https://youtu.be/lLwSusjfsbs}. It is verified that the data association results are all correct by checking the distance between the estimated vehicle position from the relative measurement and the vehicle's ground truth position. If the distance is smaller than a threshold (half of the vehicle size), the data association result is correct, otherwise, two vehicles would be in collision which does not happen in the simulation. The obtained spatial relative pose information are then fed into a pose graph for jointly pose estimation.

\subsubsection{Accuracy Evaluation}
\begin{figure}[!htb]
	\centering
	\includegraphics[width=3.5in]{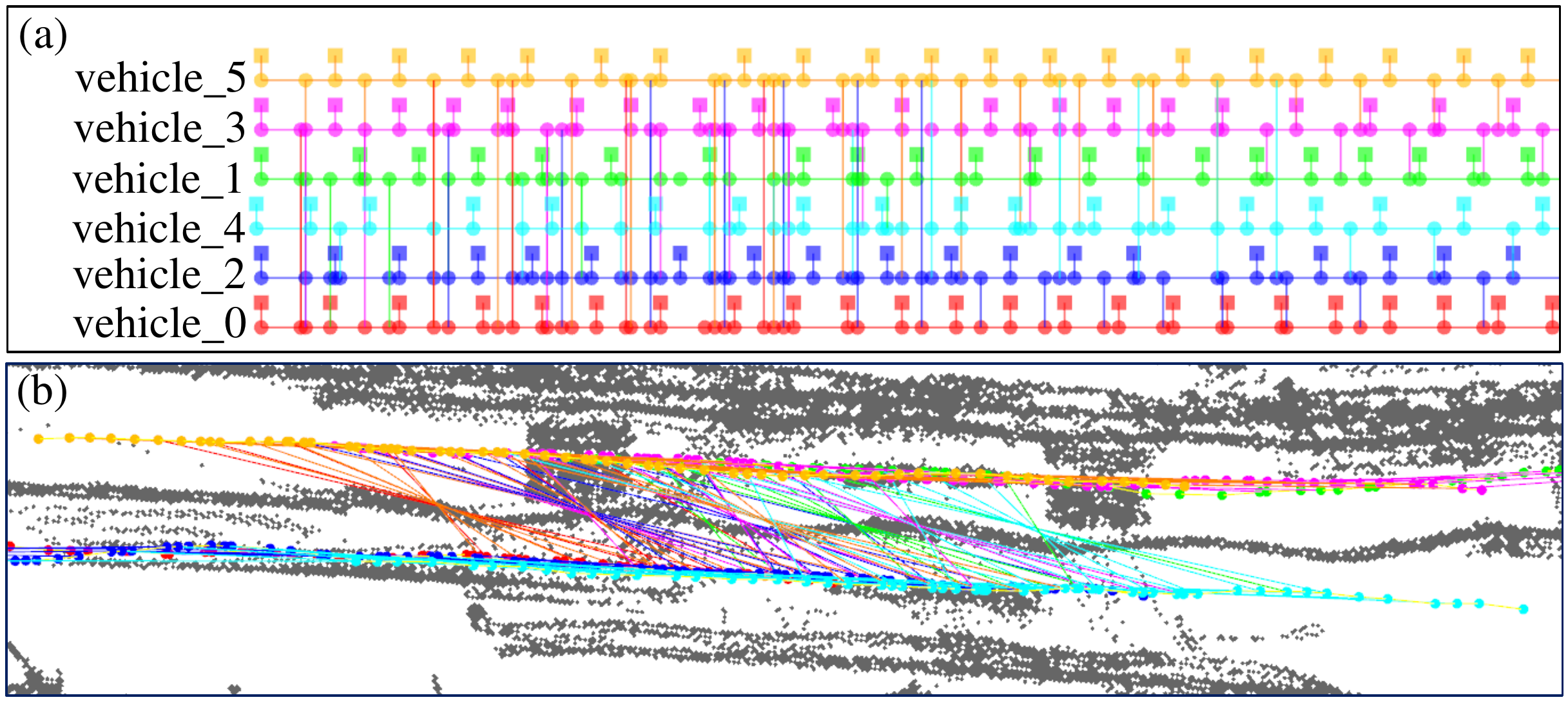}
	\caption[The pose graph for multi-vehicle cooperative localization on a straight road.]{The pose graph for multi-vehicle cooperative localization on a straight road. The pose nodes are visualized along the timeline in (a), are visualized at the estimated spatial position in (b). The circles represent the pose nodes and the squares represent the map measurements. The lines between two pose nodes of the same color represent temporal relative pose information, i.e., the horizontal lines in (a). The lines between two nodes of different colors represent spatial relative pose information, i.e., the vertical lines between two circles. The color of an edge of spatial relative pose information represents the observer vehicle's color.}
	\label{fig_pose_graph_multi_vehicle_on_a_straight_road}
\end{figure}

\begin{figure}[!htb]
	\centering
	\includegraphics[width=3.5in]{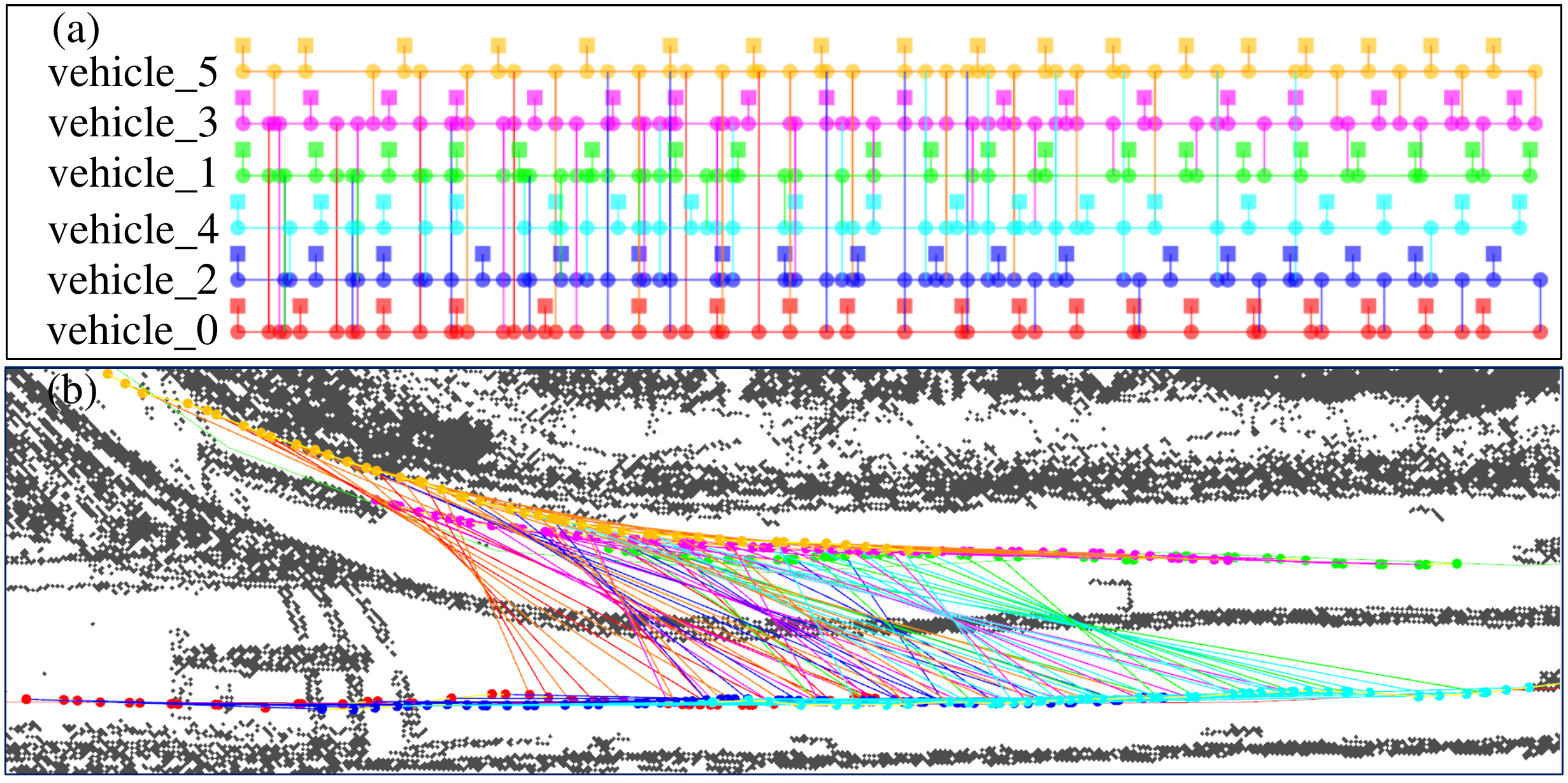}
	\caption[The pose graph for multi-vehicle cooperative localization on a curvy road.]{The pose graph for multi-vehicle cooperative localization on a curvy road. The pose nodes are visualized along the timeline in (a), are visualized at the estimated spatial position in (b).}
	\label{fig_pose_graph_multi_vehicle_on_a_curvy_road}
\end{figure}

The map measurements, temporal relative pose information, and spatial relative pose information are added into a pose graph for cooperative localization, where iSAM \cite{iSAM2008Kaess} is used for graph optimization. The built pose graphs for six simulated vehicles cooperatively localizing on a straight road and on a curvy road are shown in Fig.\ \ref{fig_pose_graph_multi_vehicle_on_a_straight_road} and Fig.\ \ref{fig_pose_graph_multi_vehicle_on_a_curvy_road}, respectively. The pose nodes are visualized along the timeline in Fig.\ \ref{fig_pose_graph_multi_vehicle_on_a_straight_road}(a), are visualized at the estimated spatial position in Fig.\ \ref{fig_pose_graph_multi_vehicle_on_a_straight_road}(b). The circles represent the pose nodes and the squares represent the map measurements. The lines between two pose nodes of the same color represent temporal relative pose information, i.e., the horizontal lines in Fig.\ \ref{fig_pose_graph_multi_vehicle_on_a_straight_road}(a). The lines between two nodes of different colors represent spatial relative pose information, i.e., the vertical lines between two circles. The color of an edge of spatial relative pose information represents the observer vehicle's color. A video showing the pose graph construction can be found at \url{https://youtu.be/rfa9od4g-0Y}.

The statistics of the errors of independent localization (IL) and cooperative localization (CL) are shown in Table \ref{table_CL_simulation_straight_road_result} and Table \ref{table_CL_simulation_curvy_road_result}. The error is computed by comparing the difference between the estimated poses and the ground truth poses in the simulation. In these two tables, ``Ave." and ``Std." represent the average and standard deviation of the estimation error, respectively. The essential difference between independent and cooperative localization is that the spatial relative pose information is utilized in cooperative localization. As we can see from Table \ref{table_CL_simulation_straight_road_result}, when vehicles are moving on the straight road as shown in Fig.\ \ref{fig_pose_graph_multi_vehicle_on_a_straight_road}, the average position error of CL is about $0.08$ m smaller than that of IL. The average orientation error of CL is about $0.44$ degree smaller than that of IL. From Table \ref{table_CL_simulation_curvy_road_result}, when vehicles are moving on the curvy road as shown in Fig.\ \ref{fig_pose_graph_multi_vehicle_on_a_straight_road}, the average position error of CL is about $0.11$ m smaller than that of IL. The average orientation error of CL is about $0.84$ degree smaller than that of IL. The accuracy improvement justifies the effectiveness of the proposed cooperative localization algorithm.

\begin{table}[!htb]
	\renewcommand{\arraystretch}{1.2}
	\caption{Error of localization on a straight road} \label{table_CL_simulation_straight_road_result} \centering
	\begin{tabularx}{3.5in}{|c|c|c|X|X|X|X|X|X|}
		\hline
		\multicolumn{3}{|c|}{Vehicle ID} & 0 & 1 & 2 & 3 & 4 & 5  \\
		\hline
		\multirow{6}{*}{CL} & \multirow{3}{*}{Position (m)} & Ave. & 0.19 & 0.17 & 0.17 & 0.16 & 0.18 & 0.16 \\ \cline{3-9}
		& & Std. & 0.09 & 0.11 & 0.09 & 0.07 & 0.15 & 0.08
		\\ \cline{2-9}
		& \multirow{3}{*}{Orientation (deg)} &  Ave.  & 0.73 & 0.64 & 0.46 & 0.67 & 0.62 & 0.55
		\\ \cline{3-9}
		& & Std. & 0.44 & 0.54 & 0.45 & 0.41 & 0.64 & 0.46
		\\ 
		\hline
		\multirow{6}{*}{IL} & \multirow{3}{*}{Position (m)} & Ave. & 0.26 & 0.29 & 0.24 & 0.27 & 0.24 & 0.32
		\\ \cline{3-9}
		& & Std. & 0.15 & 0.14 & 0.11 & 0.15 & 0.13 & 0.15
		\\ \cline{2-9}
		& \multirow{3}{*}{Orientation (deg)} &  Ave.  & 1.03 & 1.02 & 0.93 & 1.09 & 0.87 & 1.38
		\\ \cline{3-9}
		& & Std. & 0.84 & 0.80 & 0.68 & 0.94 & 0.72 & 1.01
		\\ 
		\hline
	\end{tabularx}
\end{table}

\begin{table}[!htb]
	\renewcommand{\arraystretch}{1.2}
	\caption{Error of localization on a curvy road} \label{table_CL_simulation_curvy_road_result} \centering
	\begin{tabularx}{3.5in}{|c|c|c|X|X|X|X|X|X|}
		\hline
		\multicolumn{3}{|c|}{Vehicle ID} & 0 & 1 & 2 & 3 & 4 & 5  \\
		\hline
		\multirow{6}{*}{CL} & \multirow{3}{*}{Position (m)} & Ave. & 0.21 & 0.19 & 0.15 & 0.12 & 0.17 & 0.21 \\ \cline{3-9}
		& & Std. & 0.12 & 0.13 & 0.09 & 0.07 & 0.14 & 0.12
		\\ \cline{2-9}
		& \multirow{3}{*}{Orientation (deg)} &  Ave.  & 0.91 & 0.45 & 0.43 & 0.41 & 0.52 & 0.77
		\\ \cline{3-9}
		& & Std. & 0.58 & 0.44 & 0.35 & 0.35 & 0.56 & 0.94
		\\ 
		\hline
		\multirow{6}{*}{IL} & \multirow{3}{*}{Position (m)} & Ave. & 0.29 & 0.30 & 0.22 & 0.30 & 0.27 & 0.30
		\\ \cline{3-9}
		& & Std. & 0.16 & 0.13 & 0.13 & 0.16 & 0.10 & 0.18
		\\ \cline{2-9}
		& \multirow{3}{*}{Orientation (deg)} &  Ave.  & 1.08 & 1.5 & 0.90 & 0.89 & 0.91 & 1.41
		\\ \cline{3-9}
		& & Std. & 0.95 & 0.98 & 0.71 & 0.84 & 0.51 & 1.19
		\\ 
		\hline
	\end{tabularx}
\end{table}

\subsection{Real Experiments}
\subsubsection{Experimental Setup}
\begin{figure}[!htb]
\centering
\includegraphics[width=3.5in]{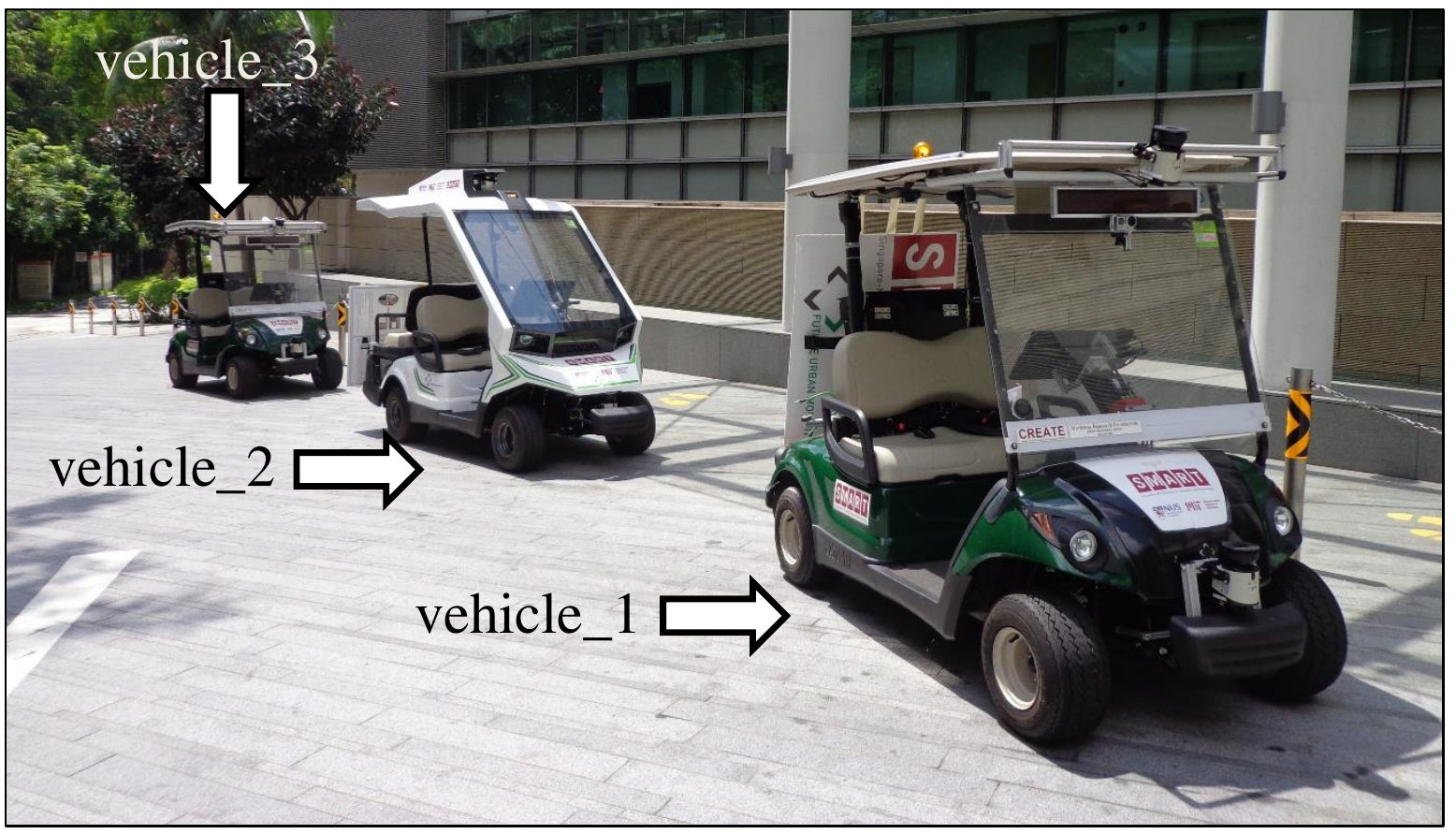}
\caption{The fleet of three test vehicles.}
\label{fig_three_buggies_on_road_experiment}
\end{figure}

Fig.\ \ref{fig_three_buggies_on_road_experiment} shows our three experimental platforms, i.e., three golfcars \cite{Multi2015Shen}, which are individually equipped with two LIDARs, an IMU, two encoders, two computers, and a Cohda wireless MK2 communication device for V2V communication. The tilt-down LIDAR is used for obtaining map measurements \cite{chong2013synthetic}. The fusion of encoder readings with IMU readings can provide odometry information, i.e., temporal relative observations. The horizontal LIDAR at the front bottom of the vehicle can detect other vehicles and an L-shape fitting algorithm \cite{EfficientL2015Shen} together with our data association algorithm is carried out to obtain spatial relative observations. All these three types of measurements are broadcast through V2V communication so that each vehicle can individually construct a pose graph based on the received information. 

\begin{figure}[!htb]
\centering
\includegraphics[width=3.5in]{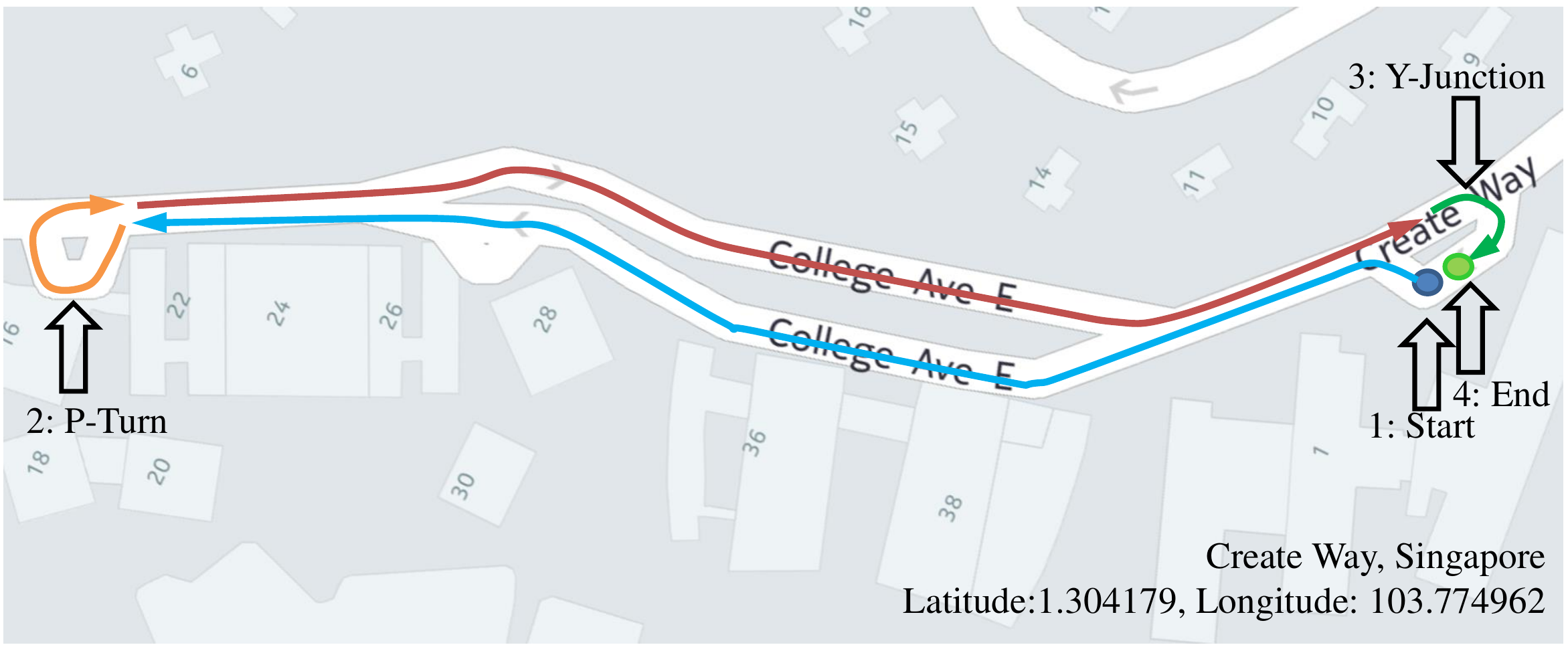}
\caption{Test road at NUS. The fleet of test vehicles move from \textit{Start} to \textit{End} via \textit{P-Turn} and \textit{Y-Junction}.}
\label{fig_experiment_road}
\end{figure}

The experiment was performed on a typical urban road in National University of Singapore (NUS). The test road is shown in Fig.\ \ref{fig_experiment_road}, which consists of Y-Junctions, a P-Turn, curvy segments, straight segments, and slopes.

\subsubsection{Precision Evaluation}
\begin{figure}[!htb]
	\centering
	\includegraphics[width=3.5in]{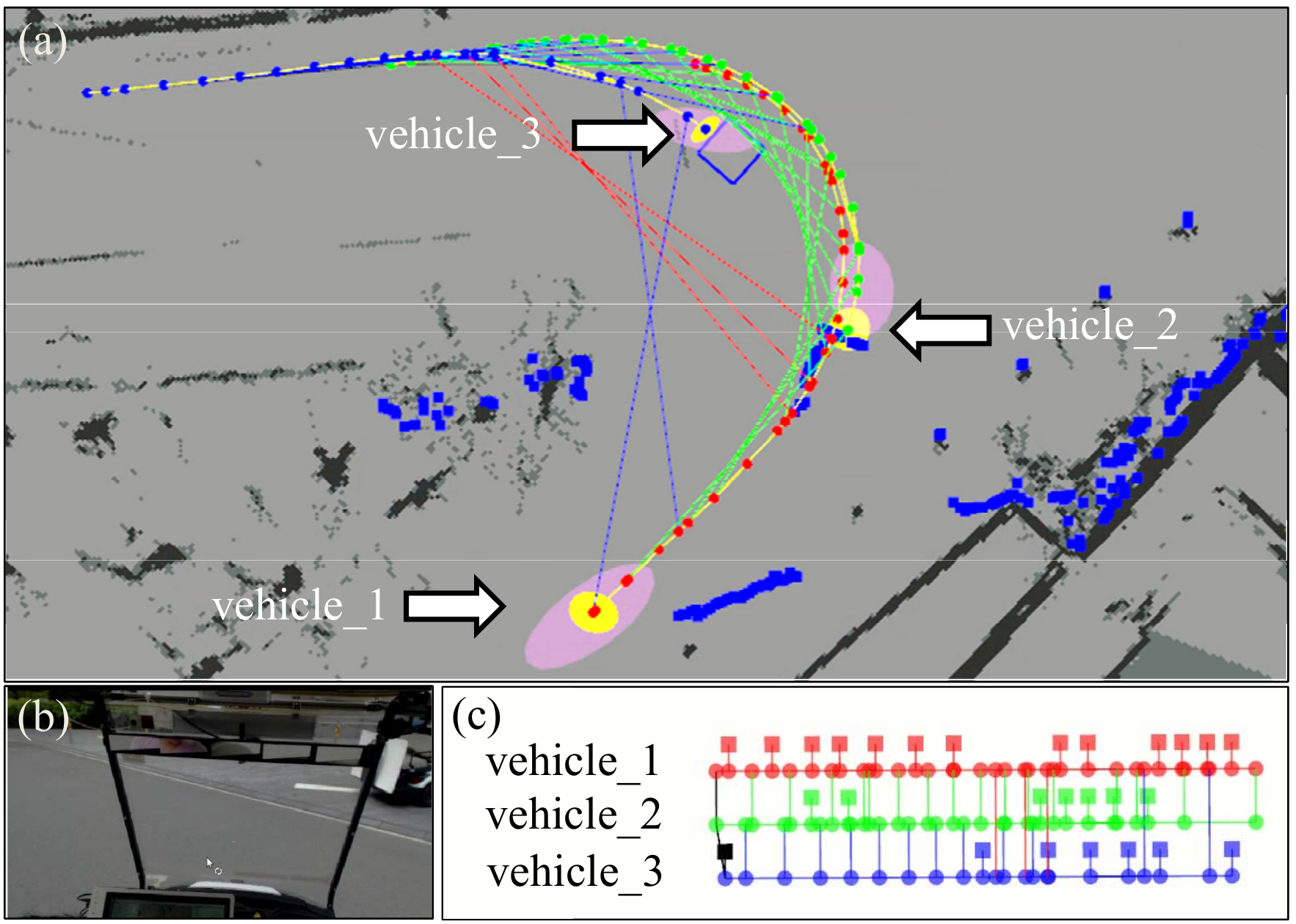}
	\caption{A snapshot of the constructed pose graph by vehicle\_3 at the \textit{Y-Junction}. The pose graph is plotted according to the spatial location in (a) and according to the timeline in (c). The black lines and squares in (a) and (c) represent the produced generic linear constraints (GLC) by the generic node removal algorithm \cite{carlevaris2014generic} when performing graph marginalization. The pink and yellow ellipses represent the location uncertainty using Independent Localization (IL) method and our Cooperative Localization method, respectively. The corresponding image from the on-board camera of vehicle\_3 is shown in (b).}
	\label{fig_YJunctionSnapshot}
\end{figure}

Since the ground truth vehicle poses are unknown in real experiments, we only examine how the pose estimation uncertainty can be reduced with our cooperative localization algorithm. In the experiment, the measurements within 10 seconds time window are all kept in the the pose graph. The measurements beyond this time window are either marginalized out using the generic node removal algorithm \cite{carlevaris2014generic} or ignored. In this way, our algorithm can accommodate up to 10 seconds delayed measurements. Fig.\ \ref{fig_YJunctionSnapshot} shows the estimated localization uncertainty (by vehicle\_3) is reduced with our pose graph based cooperative localization (CL) algorithm. The yellow ellipses shows the location uncertainty (three standard deviations) using our cooperative localization algorithm, while the pink ones shows that using independent localization (IL) algorithm \cite{chong2013synthetic}. It can be seen that the yellow ellipses are much smaller than pink ones, which indicates that the localization uncertainty is reduced.

\begin{figure}[!htb]
	\centering
	\includegraphics[width=3.5in]{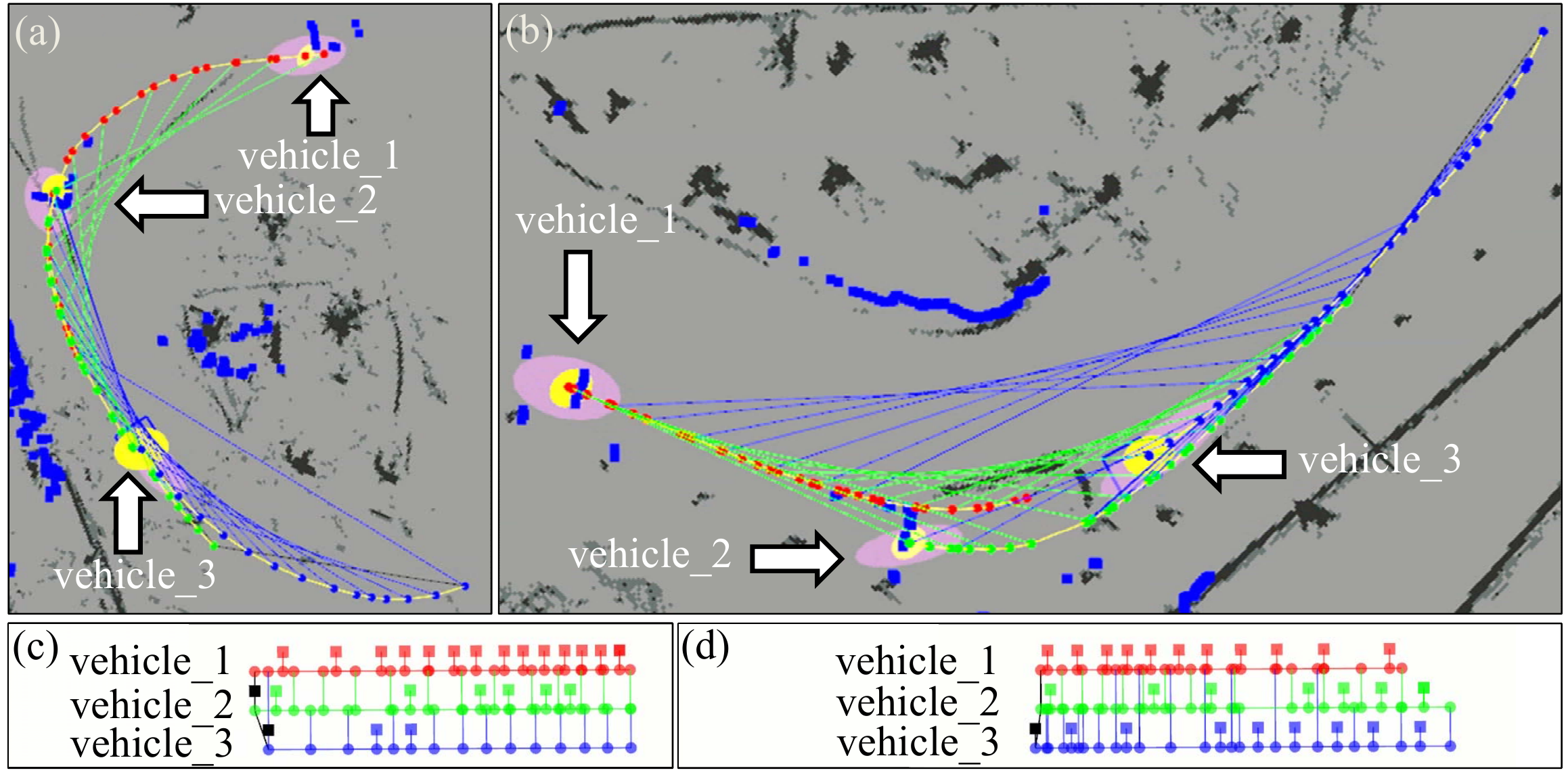}
	\caption{Two snapshots of the constructed pose graph by vehicle\_3 taken at the \textit{P-Turn} (a) and a curvy road segment near the \textit{End} (b), respectively. The corresponding pose graphs plotted according to the timeline are shown in (c) and (d). The black lines and squares represent the produced generic linear constraints (GLC) by the generic node removal algorithm \cite{carlevaris2014generic} when performing graph marginalization.}
	\label{fig_PTurnCurvyRoadSnapshot}
\end{figure}

It should be highlighted that at the \textit{Y-Junction}, as shown in Fig.\ \ref{fig_YJunctionSnapshot}, vehice\_1 can sometimes observe vehicle\_3 and vice versa, which further correlates the vehicle poses and increases the localization precision. Even though the pose graph contains many loops, as shown in Fig.\ \ref{fig_YJunctionSnapshot}(c), our algorithm avoids the overestimate problem (circular inference) and produces the optimal estimates since the measurement dependencies are described in the pose graph. Two other snapshots of the pose graph, at the \textit{P-Turn} and at a curvy road segment, are shown in Fig.\ \ref{fig_PTurnCurvyRoadSnapshot}. As we can see in Fig.\ \ref{fig_PTurnCurvyRoadSnapshot}, the spatial relative observations strongly correlates the vehicle poses, which are all incorporated into a pose graph for optimal pose estimation. The cooperative localization algorithm was running in real-time and a video showing vehicle\_3's pose graph construction can be found at \href{https://youtu.be/Dvs109UAzZ0}{https://youtu.be/Dvs109UAzZ0}. The localization uncertainty of vehicle poses are recorded during the experiment and the statistics are shown in Table \ref{table_CL_localization_uncertainty}. As we can see, compared with IL algorithm, the position uncertainty is further reduced about 0.46 meters and the orientation uncertainty is further reduced about 2.9 degrees with our CL algorithm. The localization uncertainty reduction indicates that our cooperative localization increase the precision of pose estimation.

\begin{table}[!htb]
	\renewcommand{\arraystretch}{1.2}
	\caption{Localization Uncertainty of Three Vehicles} \label{table_CL_localization_uncertainty} \centering
	\begin{tabularx}{3.5in}{|c|c|c|X|X|X|}
		\hline
		\multicolumn{3}{|c|}{Vehicle ID} & 1 & 2 & 3  \\
		\hline
		\multirow{6}{*}{CL} & \multirow{3}{*}{Position (m)} & Ave. & 0.33 & 0.36 & 0.41 \\ \cline{3-6}
		& & Std. & 0.09 & 0.07 & 0.13
		\\ \cline{2-6}
		& \multirow{3}{*}{Orientation (deg)} &  Ave.  & 2.77 & 3.26 & 3.85
		\\ \cline{3-6}
		& & Std. & 1.83 & 1.12 & 1.71
		\\ 
		\hline
		\multirow{6}{*}{IL} & \multirow{3}{*}{Position (m)} & Ave. & 0.80 & 0.89 & 0.78 
		\\ \cline{3-6}
		& & Std. & 0.17 & 0.23 & 0.15
		\\ \cline{2-6}
		& \multirow{3}{*}{Orientation (deg)} &  Ave.  & 6.24 & 6.50 & 5.87
		\\ \cline{3-6}
		& & Std. & 1.40 & 1.53 & 1.14
		\\ 
		\hline
	\end{tabularx}
\end{table}

\subsubsection{Communication Delay}
\begin{figure}[!htb]
	\centering
	\includegraphics[width=3.5in]{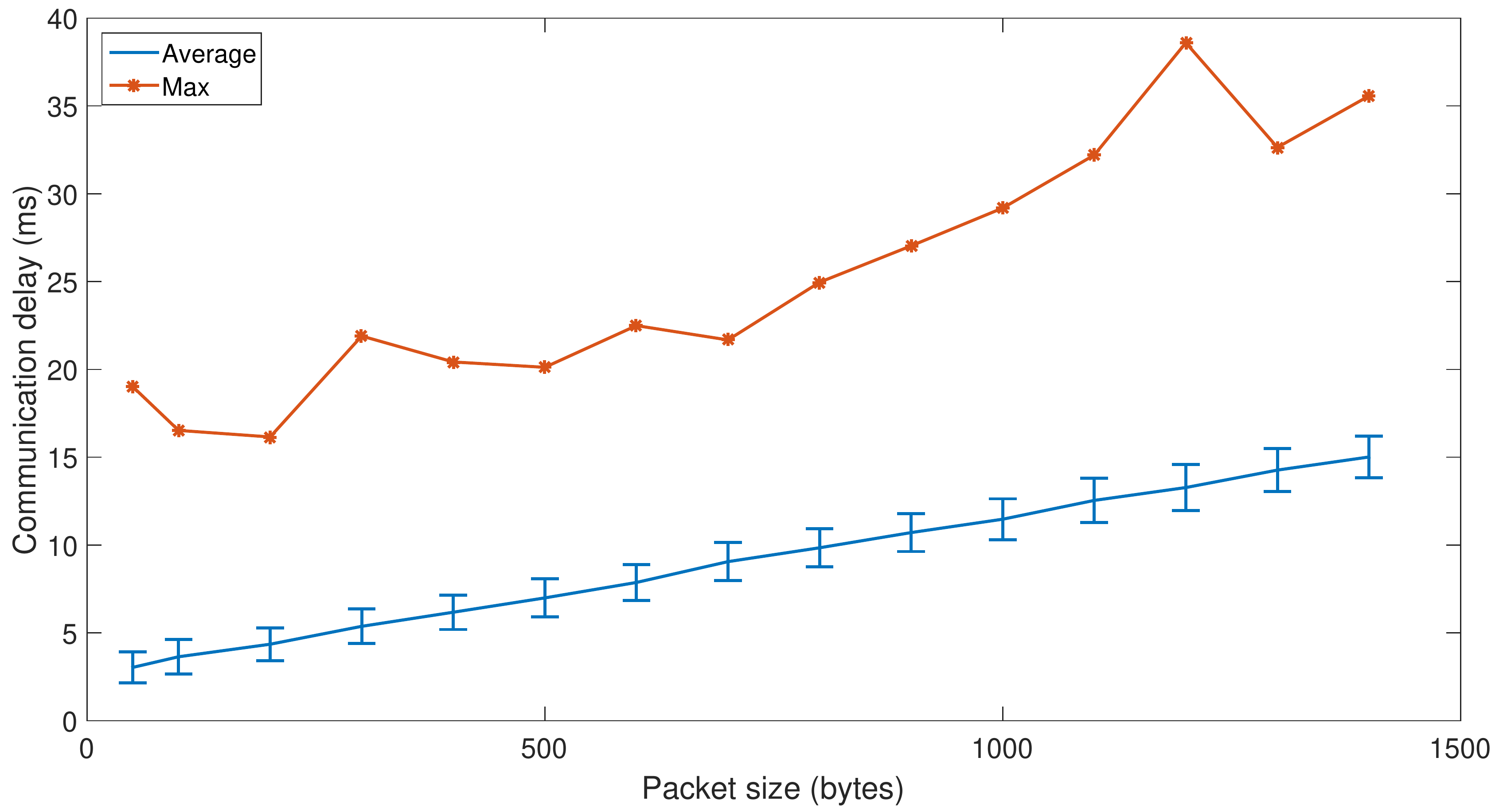}
	\caption{The average and maximum communication delay between two vehicles equipped with Cohda Wireless MK2.}
	\label{fig_communication_delay}
\end{figure}

The communication delay was measured when two vehicles (equipped with Cohda Wireless MK2) are about 15 meters away, which is about the same gap distance between vehicles in the on-road experiments. The packets are broadcast at a frequency of 10 Hz, which is also the same frequency we used in the experiments. For a particular packet size, we send 8000 packets to measure the average and the maximum of the communication delay. The result is shown in Fig.\ \ref{fig_communication_delay}.

The packet sizes for three different message types are summarized in Table \ref{table_message_packet_size}. As we can see, the packet sizes of all the messages are all below 500 bytes and thus the communication delay should be less than the maximum delay, which is 39 ms. As our time window of the pose graph is 10 seconds, which is more than sufficient to account for the delayed messages. The low latency in the communication also ensures the shared information is fused in the pose graph in a fast manner.
\begin{table}[!htb]
	\renewcommand{\arraystretch}{1.2}
	\caption{Packet Size of Different Messages} \label{table_message_packet_size} \centering
	\begin{tabularx}{3.5in}{|X|c|}
		\hline
		Message Type & Packet Size (Byte) \\
		\hline
		\hline
		Map Measurement & 409 \\
		\hline
	    Temporal Relative Observation & 198 \\
	    \hline
	    Spatial Relative Observation (0 detected vehicle) & 57 \\
	    \hline
	    Spatial Relative Observation (1 detected vehicle) & 268 \\
	    \hline
	    Spatial Relative Observation (2 detected vehicles) & 479 \\
	    \hline
	\end{tabularx}
\end{table}

In this section, both simulation and experimental results show that our cooperative localization using pose graph increased the localization accuracy and precision. The V2V communication delay and packet sizes were measured to show that our general framework can be used in practical applications.

\section{Conclusion}
In this paper, we proposed a general framework for multi-vehicle cooperative localization using pose graph. In contrast to the state-of-the-art cooperative localization algorithm \cite{Cooperative2015Walls}, ours can accommodate more sensing measurements. Rather than using distinctive spectrum of communication signals for discriminating different vehicles, we formulated vehicle identification as a linear programming problem, which can be solved efficiently. The ambiguity of spatial relative observations is resolved in the data association process. The experimental results showed that our cooperative localization algorithm can increase localization accuracy and precision.

\bibliographystyle{IEEEtran}
\bibliography{./reference}

\end{document}